\documentclass[review]{elsarticle}

\usepackage{lineno,hyperref,amsmath,amssymb,graphicx,subfig}
\modulolinenumbers[5]

\journal{Journal of \LaTeX\ Templates}

\begin{document}

\begin{frontmatter}

\title{Single Image Automatic Radial Distortion Compensation Using Deep Convolutional Network}

\author{Igor Janos\fnref{mymainaddress}}

\author{Wanda Benesova\fnref{mymainaddress}}

\cortext[mycorrespondingauthor]{Corresponding author}
\ead{wanda_benesova@stuba.sk}

\address[mymainaddress]{Slovak Technical University, Faculty of Informatics and Information Technologies}

\begin{abstract}
In many computer vision domains, the input images must conform with the pinhole camera model, where straight lines in the real world are projected as straight lines in the image. Performing computer vision tasks on live sports broadcast footage imposes challenging requirements where the algorithms cannot rely on a specific calibration pattern, must be able to cope with unknown and uncalibrated cameras, radial distortion originating from complex television lenses, few visual clues to compensate distortion by, and the necessity for real-time performance. We present a novel method for single-image automatic lens distortion compensation based on deep convolutional neural networks, capable of real-time performance and accuracy using two highest-order coefficients of the polynomial distortion model operating in the application domain of sports broadcast.
\end{abstract}

\begin{keyword}
\texttt 
Deep Convolutional Neural Network \sep%
Radial Distortion \sep%
Single Image Rectification
\end{keyword}
\end{frontmatter}



\section{Introduction}
Camera calibration is the process of finding the camera intrinsic parameters, mainly the focal length and the parameters of a chosen distortion model, to produce a pinhole-like image where straight lines in the real world are projected into straight image lines. There are known methods to perform camera calibration in laboratory environment using a set of proper calibration patterns, however once a camera is calibrated, its intrinsic parameters must remain constant. 
Single-image calibration is the process of finding the camera intrinsic parameters using just one image and is useful in situations when the camera intrinsic parameters might change on a frame-by-frame basis. A typical example might be a television broadcast of a sport event, where multiple broadcast-grade cameras might be used with wide-angle or zoom lenses. When following the play, every image captured by the cameras might exhibit different distortion due to the changing focal length and changing the internal configuration of the zoom lens. Sports footage is usually shot from longer distances with a narrow field of view resulting in a rather subtle radial distortion but of complex nature due to the elaborate optical system inside the zoom lens. 

\paragraph{Polynomial distortion model}
The distortion model is a mathematical relationship that allows conversion between the observed distorted image coordinates $x=(x_i,y_i)$ and the ideal pinhole coordinates $p=(x_p,y_p)$. The polynomial model \cite{brown1971-close} 
says that coordinates in the observed images are displaced away or towards the image center by an amount proportional to their radial distance. 
There is
\begin{equation} \label{e_1}
x=(1+k_1 ‖p‖^2+k_2 ‖p‖^4+k_3 ‖p‖^6+...) p
\end{equation}
where $k_1,k_2,k_3,...$are called the radial distortion parameters (or coefficients). For many applications using just one or two lowest order coefficients can achieve sufficient accuracy, however higher order coefficients might be necessary to model complex distortion effects.
To compute an inverse distortion, a Drap-Lefevre method \cite{drap2016-exact} 
now exists to compute the coefficients of the inverse distortion model directly. Alternatively, other approximation or iterative methods \cite{alvarez2009-algebraic}
 might be used as well.
 
 Wang et al. \cite{wang2009-simple} object to the use of polynomial model arguing that prohibitively large number of non-zero coefficients are required to model severe distortion, and that the polynomial model only works well for small distortion.

\paragraph{Division distortion model} 
The division model introduced by Fitzgibbon \cite{fitz2001-simult} is written as
\begin{equation} \label{e_2}
 p= \frac{1}{1+  \lambda_1 ‖x‖^2+  \lambda_2 ‖x‖^4+  \lambda_3 ‖x‖^6+...}  x
\end{equation}
where $\lambda_1,\lambda_2,\lambda_3,...$ are its parameters. It is crucial to remember that the division model is not an approximation to the polynomial model. Both are approximations to the camera’s true distortion function.

\paragraph{Geometric based methods}
Geometric based methods have been known to work well especially with the single-parameter division model. The division model is very popular because it can compensate better for strong distortion with just one parameter and is easier to work with. To compute the inverse distortion, it is only necessary to solve a second-order polynomial. For most webcams or cheap consumer-grade cameras with simple lenses, a single-parameter polynomial model or a single-parameter division model might be enough. However, to compensate for the distortion of a broadcast-grade lens, a higher-order polynomial model needs to be employed. Another reason for the need for a more precise model is the use of high-resolution standards in the television broadcast domain.
Geometric based methods work best when the distorted image contains strong lines, a typical feature of man-made objects. Many exploit the property of the single-parameter division model that maps straight lines into circles or arcs, and by finding multiple circles in the distorted image, the distortion parameter can be estimated. 

\paragraph{Our contribution}
In this work, we present a method to recover the camera distortion parameters from a single image using a deep convolutional neural network, which is robust to images containing only subtle clues such as the shades and patterns of grass on the sports playfield. We employ a two-parameter polynomial distortion model.
We advance with respect to state of the art with three main contributions: 
\begin{itemize}
\item representation for the radial distortion that is independent of the focal length and image resolution. 
\item a single-image method based on deep convolutional neural networks for direct estimation of $k_1$ and $k_2$ distortion model parameters applicable in the domain of sports broadcast. 
\item distortion loss function that treats the $k_1$ and $k_2$ parameters as the means of achieving a single goal instead as two independent features.
\end{itemize}


\section{Related work}

Several methods have been proposed to compensate for the lens distortion. Methods based on point correspondence \cite{brauer2004-simplemethod}, \cite{tsai1987-versatilecamera}, \cite{zhang2000-flexible}, multiple view autocalibration \cite{fitz2001-simult}, \cite{hartley2007-parameterfree}, and plumb line \cite{wang2009-simple}, \cite{bukhari2013-automatic}, \cite{wu2017-correction} were dominant before the era of deep neural networks. While point correspondence and multiple view autocalibration methods require known patterns or sequence of images under camera motion, the plumb line methods estimate the distortion parameters directly from the distorted straight lines in one or more images.

\paragraph{Pre-neural network era}

Methods based on a single parameter division model exploit the fact, that straight lines are mapped as circular arcs \cite{wang2009-simple}, \cite{bukhari2013-automatic}, \cite{wu2017-correction}. If at least three different circular arcs from three different straight lines are located in the image, the Wang method \cite{wang2009-simple} can also estimate the center of the radial distortion. Multiple optimization techniques of circle fitting have been proposed by the authors, recommending the Levenberg-Marquardt scheme as the most promising. 
Later, Bukhari and Dailey \cite{bukhari2013-automatic} have improved the Wang method \cite{wang2009-simple} by designing a two-step random sampling process to make the circle fitting fully automatic and robust to outliers. In the first step of their algorithm, they introduced a sampling algorithm to search the input image for subsequences of contour pixels that could be modeled as circular arcs. The sampling algorithm was inspired by RANSAC, but, rather than finding a single model for all the data, the algorithm preserved all the circular arc candidates not overlapping with other arcs in the same contour that had more support. After all the arc candidates were identified, a refinement procedure was applied using the inlier pixel contour subsegment supporting all the given candidates. Finally, the circle parameters were estimated using a range of circle fitting methods, with Ransac-Pratt method found as the most promising. In the second step, they introduced a sampling algorithm that finds the distortion parameters consistent with the largest number of arcs identified by the first step, favoring the support of the longer arc candidates over the short ones.
Aleman-Flores proposed \cite{aleman2013wide} a modification of the Hough transform that incorporates the distortion parameter in the Hough-space and makes it possible to detect distorted lines even when they are not observed as continuous arcs. The modified Hough space \cite{aleman2013wide} detection would not only provide the orientation and location of the lines, but also an initial estimate of the distortion parameter.
Other methods based on modified Hough transform \cite{santana2016-iterative} can estimate the parameters of a two-parameter division model. The common trait of these methods is that they require a larger number of strong distorted lines to provide strong support for the estimated parameters. The Devernay-Faugeras \cite{devernay2001-straight} method is very robust but requires a complex polygonal approximation of the distorted lines.

\paragraph{Deep convolutional neural networks}

Rong et al. showed \cite{rong2017-radial} that training a convolutional neural network is an effective approach for radial distortion correction. Their method was based on the single parameter division model and used a classification approach to estimate the distortion parameter with a finite precision as one of 401 possible discrete values. The training data was synthesized from a subset of the ImageNet dataset that satisfied the condition of containing strong line segments detectable by the Hough transform.

López-Antequera et al. \cite{lopez2019-deepsingle} have successfully trained the convolutional neural network to jointly estimate the camera orientation (tilt, roll) and intrinsic parameters (focal length and radial distortion) from single images. Their method used synthetic data generated from SUN360 \cite{xiao2012-recognizing} panorama pictures containing man-made objects of everyday life. The method estimated two distortion parameters of the polynomial model, expressing the $k_2$ parameter as a function of $k_1$. The convolutional neural network used a DenseNet feature extractor pretrained on the ImageNet dataset. However, the method was only trained to correct barrel distortion ($k_{1}\in [-0.4;0]$). For our purposes, the pretrained feature extractor turned out to be insufficient since the football field contains features different to those contained in the ImageNet dataset. To achieve better accuracy for images of the football field we propose training a task-specific feature extractor from scratch. In this work, we also seek to find a better estimate of $k_2$ for real lenses with characteristics outside the distortion parameter manifold discovered in \cite{lopez2019-deepsingle}.

Xue et al. \cite{xue2020-fisheye} have extended the Rong \cite{rong2017-radial} framework by adding a distorted line perception (DLP) module that outputs a heat-map of significant distorted lines presented in fisheye images, and by adding a multi-scale calibration module. The Xue method \cite{xue2020-fisheye} was specifically designed to work with fisheye lenses and fisheye distortion model. The major difference between the regular lenses modeled by the low-order polynomial or division models and fisheye lenses is, that the distortion approaches infinity at the image borders. Such distortion is difficult to model using low-order polynomials. After the significant distorted straight lines are identified by the DLP module, the Multi-Scale Calibration module estimates the distortion parameters by computing first the global estimate and local estimate for the central region, and four surrounding regions (the upper left, the lower left, the upper right and the lower right). The authors argue that using the significant lines heat-map makes the network learn to estimate the distortion parameters faster.

Liao et al. \cite{liao2021-deepordinal} have extended the Rong \cite{rong2017-radial} framework by adding a new representation for the lens distortion, called the Ordinal Distortion. Instead of estimating the significant lines, the Liao method \cite{liao2021-deepordinal} tries to estimate the distortion rate for image features with gradually increasing distance from the image center, the shape of the distortion function sampled at given radii. The authors argue that the distortion information is redundant in the distorted images and exhibits central and mirror symmetry with respect to the principal point.

\section{Theoretical background}

In this section, we describe the theoretical foundations on which our method is built.

\paragraph{Camera model}
We assume a perspective projection camera model with square pixels and a principal point located at the center of the image sensor, as described in \cite{klette1998-computervision}. The perspective projection projects a 3D point $p_{3d}=(X,Y,Z)$ into a 2D point on a plane located at $Z=1$ as normalized image coordinates $p=(x,y)= (X/Z,Y/Z)$. The normalized image coordinates are scaled by a factor $d$ corresponding to the radial distortion, that can be expressed as:

\begin{equation} \label{e_3}
    r=\sqrt{x^2 + y^2}
\end{equation}
\begin{equation} \label{e_4}
    d=1 + k_1r^2 + k_2r^4
\end{equation}
\begin{equation} \label{e_5}
    p_d = (x_d, y_d) = (dx, dy)
\end{equation}
\begin{equation} \label{e_6}
    r_d = dr
\end{equation}

The distortion factor is a function of the radius $r$ and distortion model parameters $k_1$,$k_2$. Scaling the distorted normalized image coordinates by the focal length $f$ yields the resulting image pixel coordinates $p_{di}=(u_d,v_d)=(fx_d,fy_d)$, relative to the image sensor center. Applying the scale factor $f$ to the radii of normalized and distorted image coordinates, the following equations are true:

\begin{equation} \label{e_7}
\frac{r_i}{f}=r
\end{equation}
\begin{equation} \label{e_8}
\frac{r_{di}}{f}=r_d
\end{equation}

\paragraph{Distortion model}

To express the distortion, we consider a two-parameter polynomial distortion model. As described in the previous section, it is worth mentioning that $k_1$,$k_2$ components of the distortion factor d are related to the camera lens and remain constant (unless the internal configuration of the optical system is changed) even when the focal length is changed. In effect, the curvature of lines in the distorted image seems to be decreasing as the focal length increases because the final image contains a smaller crop of the viewing frustum.

Since scaling by focal length is a linear operation, the following relation is true for both the distorted normalized image coordinates and undistorted normalized image coordinates, as well as for the distorted image coordinates and undistorted image coordinates.

\begin{equation} \label{e_9}
    d=\frac{r_{d}}{r}=\frac{r_{di}}{r_i}
\end{equation}

After making proper substitutions in the equation (4) with the terms (7), (9) it can be shown that:

\begin{equation} \label{e_10}
    r_{di}=r_i(1+\frac{k_1}{f^2}r_i^2+\frac{k_2}{f^4}r_i^4)
\end{equation}

where the distortion factor $d$ is expressed as a function of the image pixel coordinate radius $r_i$ instead of normalized coordinate radius $r$ and a new set of distortion parameters $\hat{k_1}=k_1/f^2$, and $\hat{k_2}=k_2/f^4$ instead of $k_1$, $k_2$. The distortion specified by the parameters $\hat{k_1}, \hat{k_2}$ is called the apparent distortion \cite{lopez2019-deepsingle} and is related to the image itself. The curvature of lines in images with constant $\hat{k_1}, \hat{k_2}$ remains the same regardless of the current focal length of the camera at the time the image was captured.

It is much easier for the neural network to learn to estimate the parameters $\hat{k_1}, \hat{k_2}$ than to estimate the parameters $k_1$, $k_2$ directly. It is also sufficient to work with the apparent distortion parameters to compensate for the distortion and produce a corrected image, in which straight lines are indeed straight.

\section{Proposed method}

We estimate the parameters of a resolution- and focal length-independent distortion model to compensate for the radial distortion using a deep convolutional network. Our dataset is created from a hand-made collection of real panorama images of football stadiums taken from multiple positions inside the arena, which are then cropped and distorted in an inverse fashion using arbitrary parameters.

\paragraph{Apparent distortion at scale}

Even though the use of the apparent distortion is advantageous for neural network training, there is still one issue with it, the $\hat{k_1}, \hat{k_2}$ parameters are related to the pixel dimensions of the input image. To overcome this issue, we propose an apparent distortion at scale model specified by the parameters $\tilde{k_1}, \tilde{k_2}$. A scaled distortion can be found from any originating distortion (either distorted normalized coordinates, image pixel coordinates, or any other) by applying a scale factor $s$ as follows:

\begin{equation} \label{e_11}
    r_{scaled}=sr_{original}    
\end{equation}

\begin{equation} \label{e_12}
    \tilde{k_1}=\frac{k_1}{s^2}, \tilde{k_2}=\frac{k_2}{s^4}
\end{equation}

We propose using such a scale factor $s$, that would make the image width equal to $1$. The height of the scaled image would be proportional to the aspect ratio of the original image.
The parameters of the scaled apparent distortion would be still related to the image itself and describe the same curvature of lines in images with the same parameters regardless of the focal length. Furthermore, it is possible to experiment with different network architectures and different input sizes, yielding the same-scaled results. When implementing compensation algorithms, often frameworks for GPU-based rendering are used (such as DirectX or OpenGL) that use texture coordinates ranging from $<0;1>$ interval to address image pixels. Having the results meaningful for this range makes the implementation straightforward. If there is a need for distortion parameters relevant for a specific resolution, the parameters $\tilde{k_1}, \tilde{k_2}$ can be easily rescaled by a new scale factor.

\paragraph{Dataset}

\begin{figure}
    \centering
    \includegraphics[width=8cm]{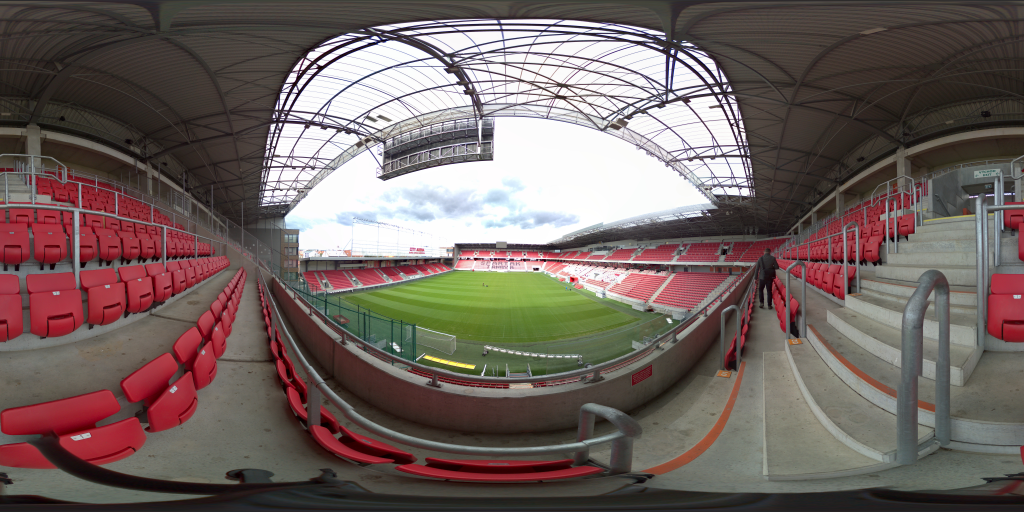}
    \includegraphics[width=8cm]{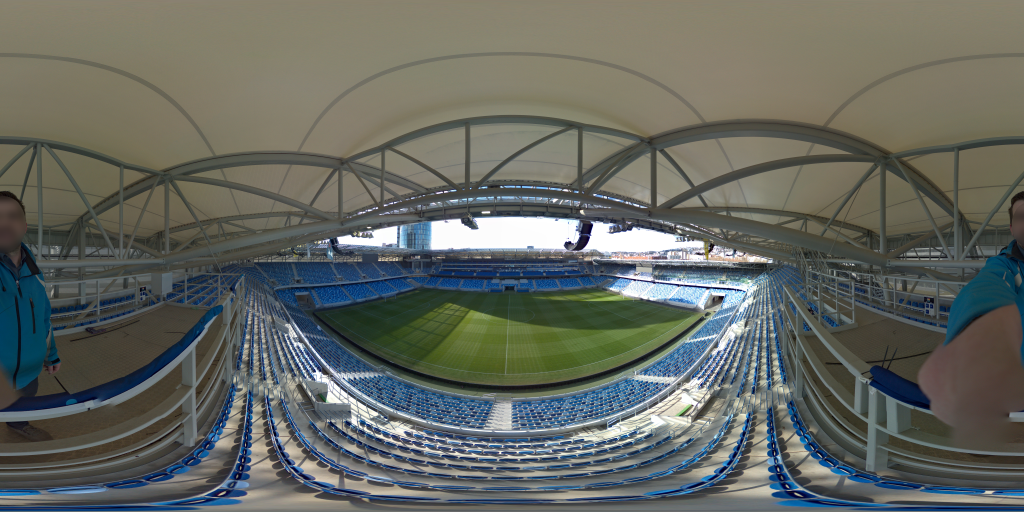}
    \caption{Example panorama images}
    \label{fig:ExampleImages}
\end{figure}

We use distorted crops of real panorama images to train our deep convolutional network. First, we obtained 218 equirectangular panorama images from several football arenas (Figure \ref{fig:ExampleImages}). In each stadium, we captured more than 50 panorama images from multiple positions in the arena’s tribune including broadcast camera platforms, each image offering a unique view of the playfield. The images contained challenging lighting conditions, bad weather, high contrast situations, football pitch maintenance situations, and a regular football match with players and referees to account for many possible situations that may happen. The equirectangular panorama images were captured in very high resolution – 16384 x 8192 pixels and adjusted to contain the football pitch at the image center.

\begin{figure}
    \centering
    \includegraphics[width=4cm]{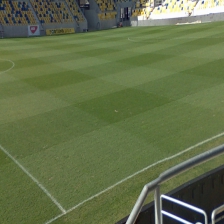}
    \includegraphics[width=4cm]{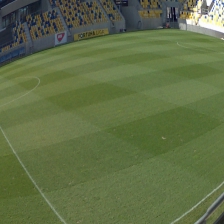}
    \includegraphics[width=4cm]{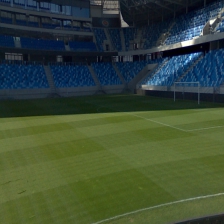}
    \includegraphics[width=4cm]{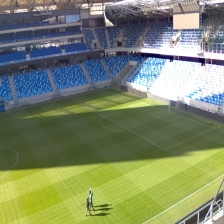}
    \includegraphics[width=4cm]{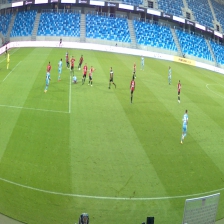}
    \includegraphics[width=4cm]{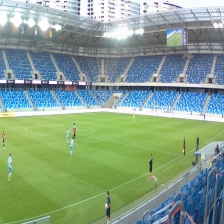}
    \caption{Example cropped images}
    \label{fig:ExampleCrops}
\end{figure}

Then, from each panorama image, we generate a high number of crop images (Figure \ref{fig:ExampleCrops}) by randomly sampling the pan, tilt, roll, field of view and distortion coefficients $\tilde{k_1}, \tilde{k_2}$. All the parameters were sampled from the probability distribution summarized in table \ref{tab:DatasetParameters}. The final dataset consisted of 2,150 million cropped images in the training set, and 10,000 cropped images in the cross-validation set, and another 10,000 images in the test set. All cropped images were first rendered at 1280x720 resolution with a 16:9 aspect ratio and then resized to 224x224 resolution to match the deep convolutional network input resolution. To generate crops from panorama images, we apply mapping from spherical image coordinates to rectilinear coordinates \cite{szeliski2010-computervision}\cite{moeslund2014computer}, which are later distorted in an inverse fashion. An inverse distortion of a 2-parameter polynomial distortion is defined by a 4-parameter $(b_1, b_2, b_3, b_4)$ polynomial, and the parameters can be found using the equations \cite{drap2016-exact}:

\begin{equation} \label{e_13}
\begin{aligned}
&b_1 = -k_1 \\
&b_2 = 3k_1^2 - k_2 \\
&b_3 = -12k_1^3 + 8k_1k_2 - k_3 \\
&b_4 = 55k_1^4 - 55k_1^2k_2 + 5k_2^2 + 10k_1k_3 - k_4
\end{aligned}
\end{equation}

Lopez-Antequera et al. \cite{lopez2019-deepsingle} have discovered that for many real cameras the $k_1$ and $k_2$ parameters seem to lie close to a one-dimensional manifold:

\begin{equation} \label{e_14}
    k_2 = 0.019k_1 + 0.805k_1^2
\end{equation}

\begin{figure}
    \centering
    \includegraphics[width=8cm]{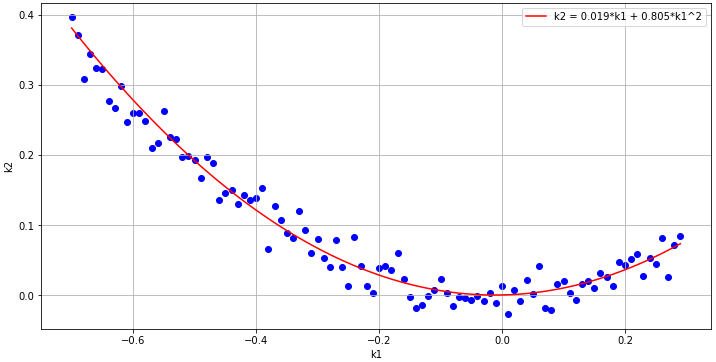}
    \caption{Distribution of $\tilde{k_1}$ and $\tilde{k_2}$ as sampled in our method}
    \label{fig:DistK1K2}
\end{figure}

In our method, we propose using a wider range for the $\tilde{k_1}$ parameters – $[-0.7; 0.3] $  to allow for stronger barrel distortion and to allow for pincushion distortion. To model the distribution of the $\tilde{k_2}$  parameters (Figure \ref{fig:DistK1K2}) of real cameras, we add a random noise factor to the $\tilde{k_2}$ values defined by the one-dimensional manifold discovered by Lopez-Antequera. We train our deep neural network to directly estimate the values of $\tilde{k_2}$.

\begin{table}[]
    \centering
    \begin{tabular}{c c c}
        Parameter & Distribution & Values \\
        \hline
        Pan & Uniform & $[-35^{\circ}; 35^{\circ}]$ \\
        Tilt & Uniform & $[-15^{\circ}; 0^{\circ}]$ \\
        Roll & Uniform & $[-2^{\circ}; 2^{\circ}]$ \\
        \\
        Field of view & Uniform & $[15^{\circ}; 60^{\circ}]$ \\
        \\
        $\tilde{k_1}$ & Uniform & $[-0.7; 0.3]$ \\
        noise of $\tilde{k_2}$ & Normal & $\mu=0.0, \sigma=0.02$ \\
        \hline
    \end{tabular}
    \caption{Distribution of the camera parameters used to generate the synthetic data set}
    \label{tab:DatasetParameters}
\end{table}

\paragraph{Network architecture}

\begin{figure}
    \centering
    \includegraphics[width=12cm]{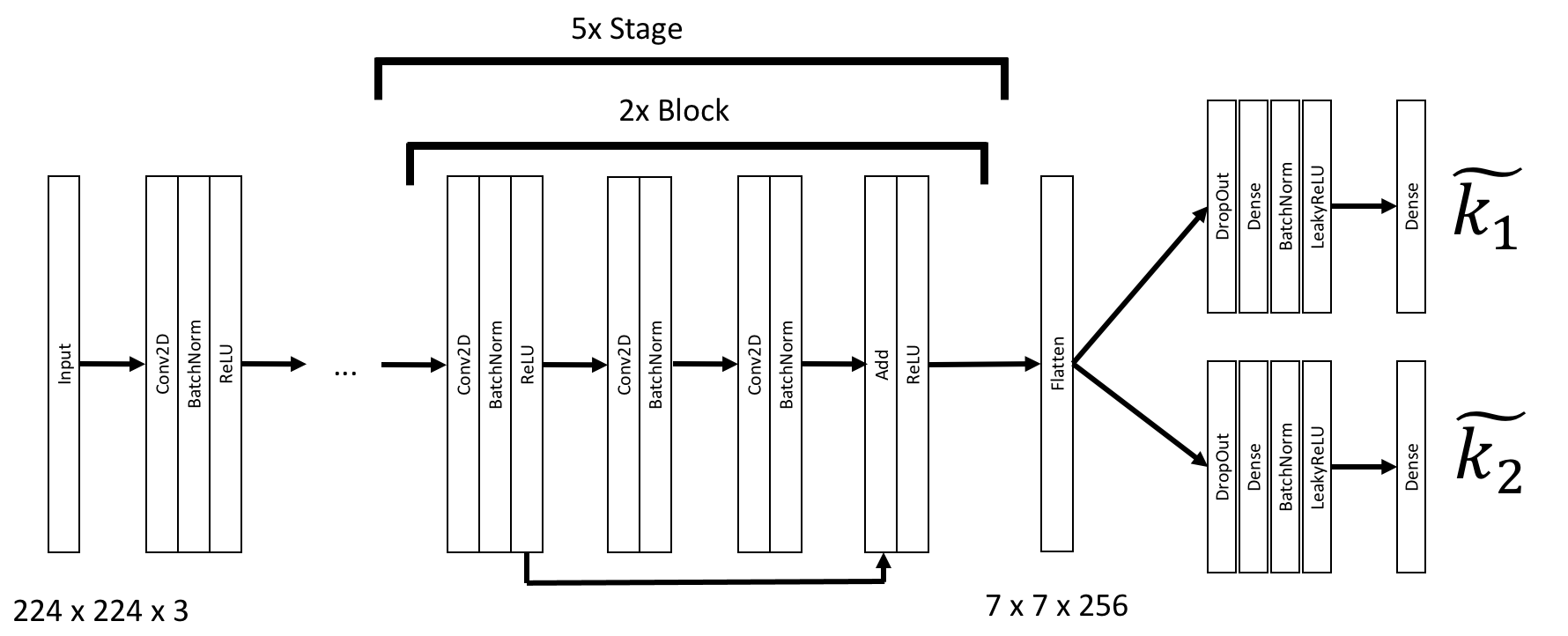}
    \caption{Network architecture}
    \label{fig:Architecture}
\end{figure}

We used a deep neural network (Figure \ref{fig:Architecture}) consisting of a feature extractor and two independent regressors outputting the values $\tilde{k_1}, \tilde{k_2}$. We have experimented with several popular architectures such as MobileNet, ResNet, or DenseNet for the feature extractor part. The more powerful architectures such as ResNet or DenseNet seemed to have overfitted the training data quickly and performed poorly on real images. We had a success with a light-weighted feature extractor inspired by the ResNet architecture with a reduced number of filters and inverted residual block repetitions. Thanks to our large, synthesized dataset, we could train the network completely from scratch and have achieved better performance compared to fine-tuning publicly available pre-trained models.

Both regressors outputting $\tilde{k_1}, \tilde{k_2}$ were built using a fully connected layer of 32 units followed by batch normalization, LeakyReLU activation, and a final single-unit fully connected layer.

\paragraph{Distortion loss function}

\begin{figure}
    \centering
    \includegraphics[width=8cm]{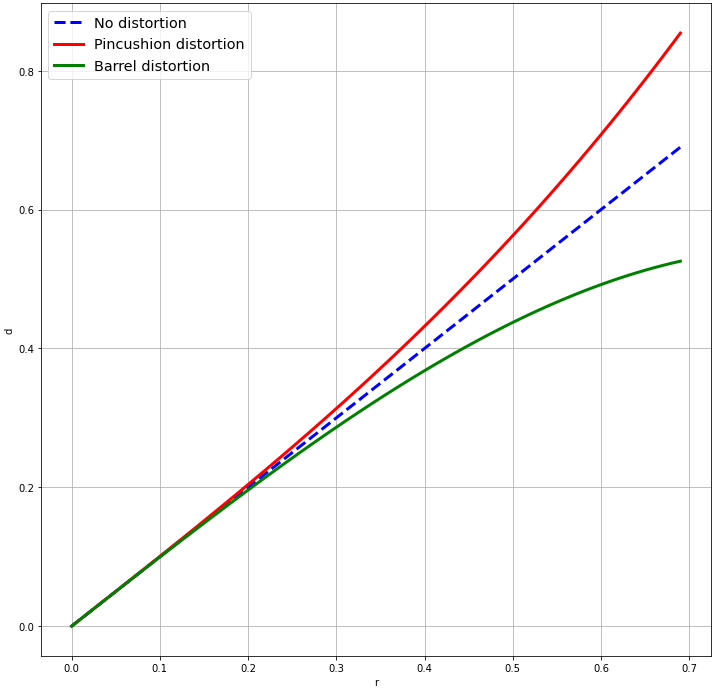}
    \caption{Basic distortion effects - pincushion, and barrel distortion}
    \label{fig:BasicDistortion}
\end{figure}

We differentiate two kinds of basic distortion effects (Figure \ref{fig:BasicDistortion}) – pincushion, and barrel. Each distortion can be thought of as a function mapping the radius (or distance) of an image pixel from the image center to a new distance value. 

As discussed in the sections above, we can define a polynomial distortion function as:

\begin{equation} \label{e_15}
    p(r, (\tilde{k_1}, \tilde{k_2})) = r(1 + \tilde{k_1}r^2 + \tilde{k_2}r^4)
\end{equation}

Depending on the values of $\tilde{k_1}, \tilde{k_2}$ the plot of the distortion function may vary, as depicted in the figure 5. 

\begin{figure}
    \centering
    \includegraphics[width=8cm]{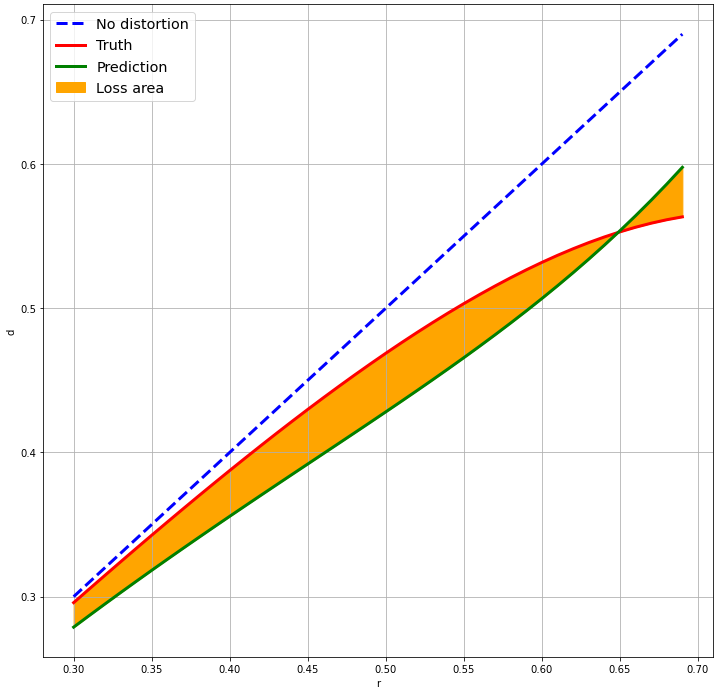}
    \caption{The plot of two distortion functions defined by two sets of $\tilde{k_1}, \tilde{k_2}$ parameters - the truth pair, and the prediction pair. Loss area between the two plots.}
    \label{fig:LossArea}
\end{figure}

We propose a distortion loss function to express the loss area (Figure \ref{fig:LossArea}) between two distortion function plots as:

\begin{equation} \label{e_16}
    \mathcal{DL}(y,\hat{y})=\sum_{i=1}^{N}(p(r_i, y) - p(r_i, \hat{y}))^2
\end{equation}

where $r_i \in \{r_1, r_2, ..., r_N \}$ are discretely finely sampled radii from an interval of $<0; 0.7>$, and $y, \hat{y}$ are true label and predicted pairs of $\tilde{k_1}, \tilde{k_2}$.
By minimizing the distortion loss function, we try to train the network to predict a pair of $\tilde{k_1}, \tilde{k_2}$ values, whose distortion function lies as close as possible to the distortion function of the ground truth coefficients.

The proposed distortion loss function can express the errors in both $\tilde{k_1}, \tilde{k_2}$ coefficients as a single measure. However, even if one of the coefficients is predicted correctly and the other is not, the error gradient will still propagate backward through the network for both coefficients making it difficult to converge on the correct solution. For this reason, we use a similar technique as described in \cite{lopez2019-deepsingle}, and evaluate the distortion loss function for each of the $\tilde{k_1}, \tilde{k_2}$ coefficients separately, and compute the final loss function as

\begin{equation} \label{e_17}
\begin{aligned}
&\mathcal{L}_{k1} = \mathcal{DL}( (\tilde{k_1}, \tilde{k_2}), (\hat{\tilde{k_1}}, \tilde{k_2}) ) \\
&\mathcal{L}_{k2} = \mathcal{DL}( (\tilde{k_1}, \tilde{k_2}), (\tilde{k_1}, \hat{\tilde{k_2}}) )
\end{aligned}
\end{equation}

\begin{equation} \label{e_18}
\mathcal{L}( (\tilde{k_1}, \tilde{k_2}), (\hat{\tilde{k_1}},\hat{\tilde{k_2}})) = \mathcal{L}_{k1} + \mathcal{L}_{k2}
\end{equation}

\section{Experiments and results}

It is rather difficult to compare the performance of our method with other related methods mentioned in chapter 2 directly, mainly because our method was trained on a dataset containing images of a football field where strong line features are very rare. In contrast, other methods specifically rely on the presence of a larger number of such strong lines. That is why we try to compare the accuracy of the distortion compensation of our two-parameter method indirectly against the two-parameter distortion model as described in \cite{lopez2019-deepsingle}, where one parameter is a function of the other.

\paragraph{Distortion reprojection error} 

First, we want to concentrate on the dataset generation process, more specifically on the part where inverse distortion is induced into the cropped images. Even though an exact algebraic solution to this problem exists, we are still using only two highest-order coefficients for distortion compensation, and only four coefficients to calculate the inverse distortion. This is inherently a non-perfect procedure. 

\begin{figure}
    \centering
    \includegraphics[width=8cm]{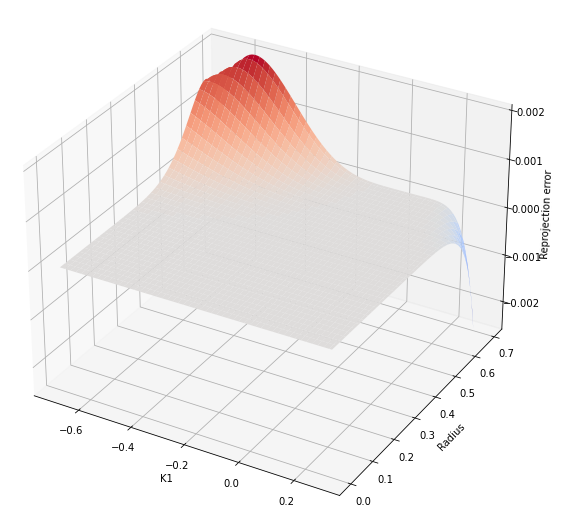}
    \caption{Reprojection error after applying the inverse and forward distortion}
    \label{fig:Reprojection}
\end{figure}

We can see in the figure \ref{fig:Reprojection}, that only in the extreme boundaries of our $\tilde{k_1}$ interval the reprojection error reaches the value of 0.002, which is roughly the equivalent of 2 pixels of a 1920 x 1080 image. For $\tilde{k_1}$ in the ranges most common for real camera lenses, the reprojection error is nearly zero.

\paragraph{Distortion function}

Second, we can compare the performance of our method to Lopez-Antequera’s method by using the distortion loss function as a metric.

We evaluate the performance of our neural network on the test set by performing:
\begin{enumerate}[(A)]
\item A direct estimate of $\tilde{k_1},\tilde{k_2}$  by the neural network as proposed by our method.
\item A direct estimate of $\tilde{k_1}$, and compute $\tilde{k_2}$ as proposed by Lopez-Antequera’s method and finally use the loss function to determine which method gives a better fit to the true distortion function.
\end{enumerate}

On our dataset, we have found out that our method of direct estimation of two distortion parameters (A) outperforms Lopez-Antequera’s method (B) in 61.2 \% of our test set images.

\section{Conclusion}

We present a method based on a deep neural network that predicts the parameters of radial lens distortion. We propose a representation for the distortion parameters independent of the camera focal length, independent of the image resolution and is easy to learn for the neural network. We have trained the network using the TensorFlow framework and GeForce RTX 3090 GPU in less than 40 hours. The light-weighted modification of the ResNet architecture has made our final network only 6.6 million parameters large, which makes it small enough to be computed efficiently even on CPUs. Our benchmarks performed on a MacBook Air with Python and OpenCV have revealed that it is able to achieve near-real-time performance on a single thread with 1920x1080 video sequences. Especially for lenses with small distortion ($\tilde{k_1}$ is close to 0), the ability to directly estimate $\tilde{k_2}$ had proven superior and resulted in a very accurate compensation of the distortion effect.

In future work, we might explore several directions of improving this method. We will look for more efficient feature extractor alternatives that might reduce the necessary network size and reduce the computations and execution time, making the method feasible for low-end edge hardware. We will try and train the network to operate on grayscale images, avoiding the need for computing color-space conversions, making it possible to process images captured by television cameras operating in YCrCb color-space directly. We will try to include short-term memory and process the input frames as a sequence to make up for images where challenging lighting conditions or the lack of information makes the method output inaccurate results. Finally, we might also include landmarks detection that would make a direct 4-point homography estimation from the distorted image possible.

\textbf{Acknowledgments:} The authors wish to thank Dušan Šott and Patrik Hrdlička for their help in acquiring the panorama images from football stadiums.

\begin{figure}
    \centering
    \includegraphics[width=12cm]{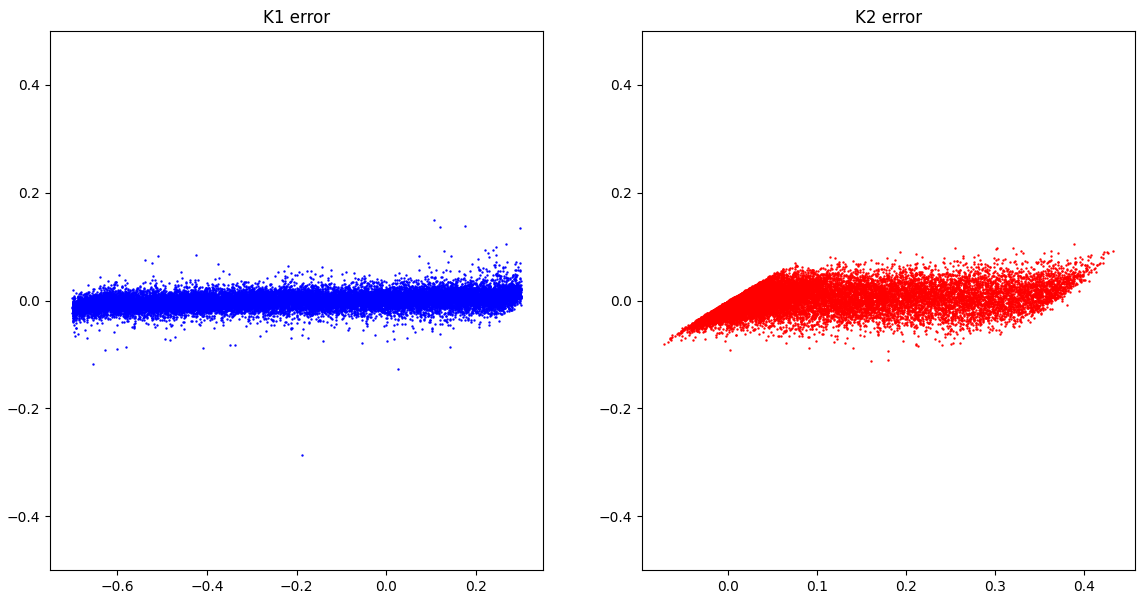}
    \caption{The final distribution of the prediction errors of $\tilde{k_1},\tilde{k_2}$ on our test set}
    \label{fig:my_label}
\end{figure}


\begin{figure}
    \centering
    \includegraphics[width=6cm]{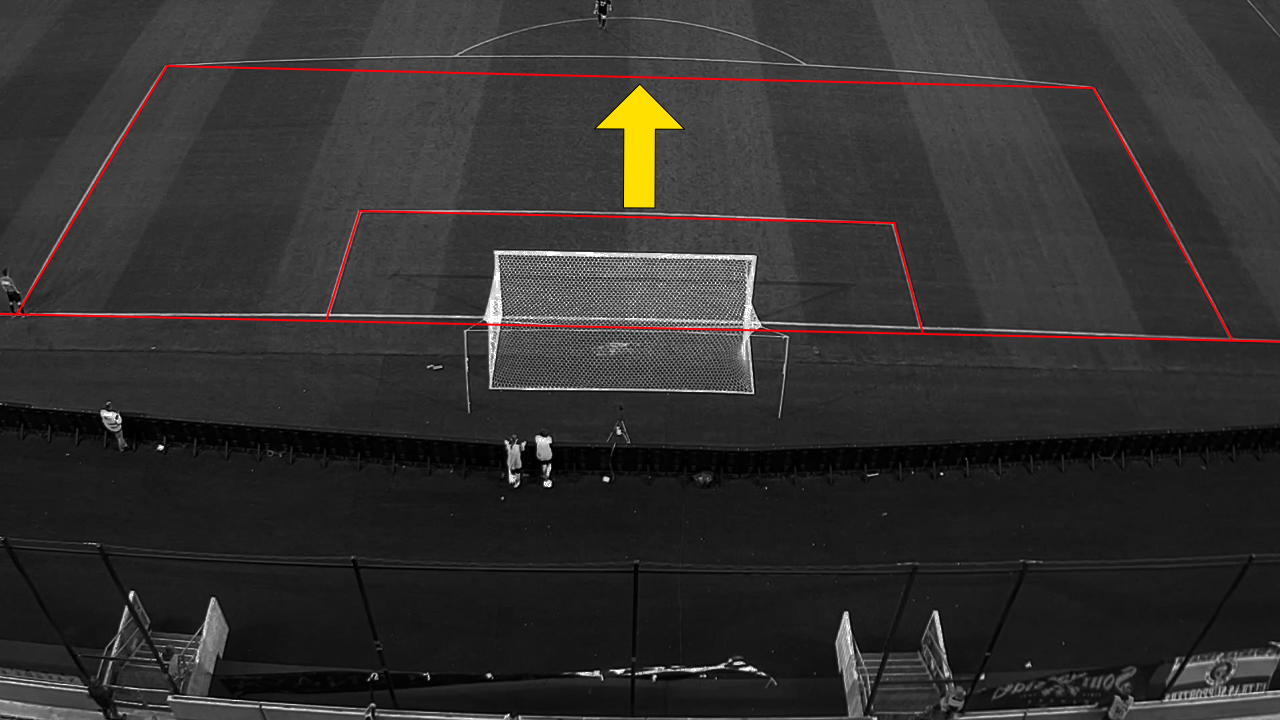}
    \includegraphics[width=6cm]{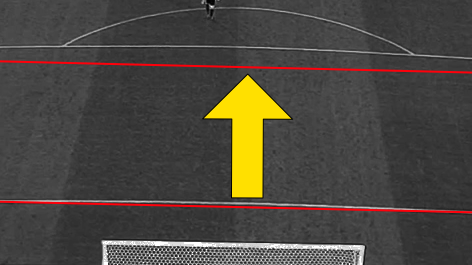}
    \includegraphics[width=6cm]{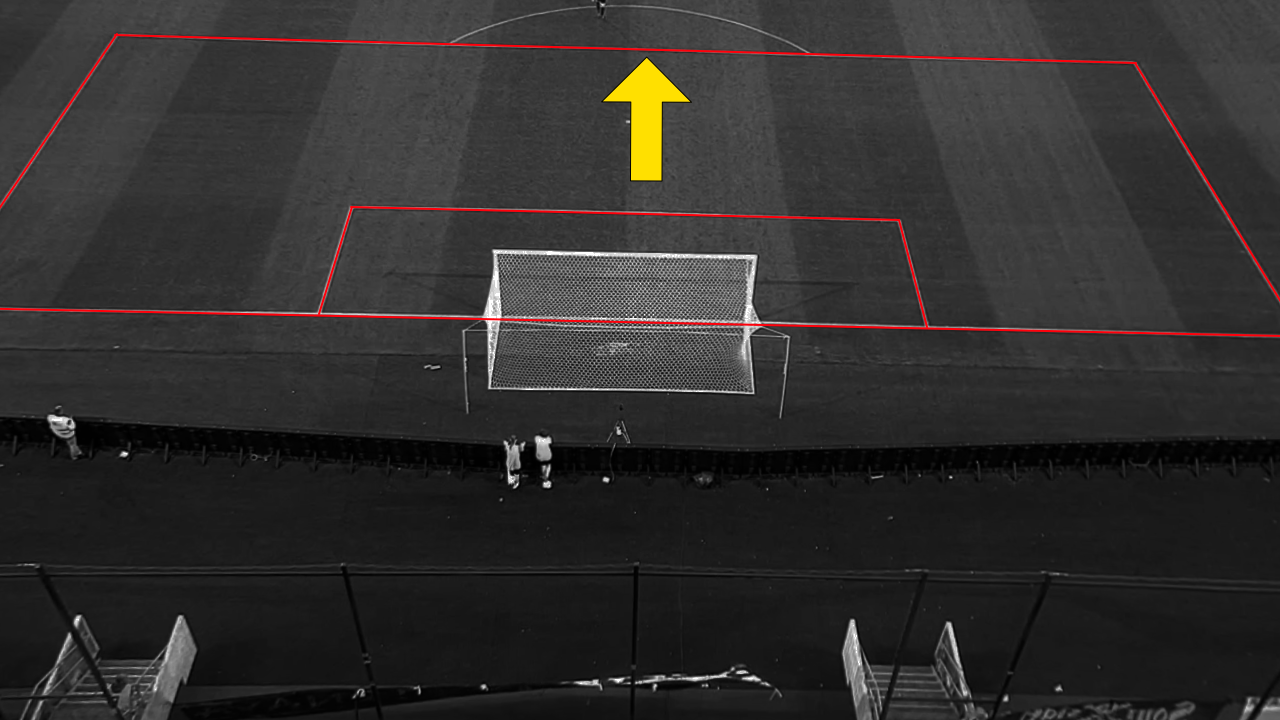}
    \includegraphics[width=6cm]{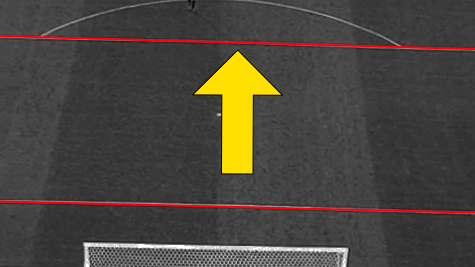}

    \caption{Original distorted image with red overlaid straight lines (top left), corrected image with red overlaid straight lines (bottom left), close-up of the distorted image where distortion is most prominent (top right), close-up of the corrected image where distortion is most prominent (bottom right).}
    \label{fig:my_label}
\end{figure}

\begin{figure}
    \centering
    \includegraphics[width=6cm]{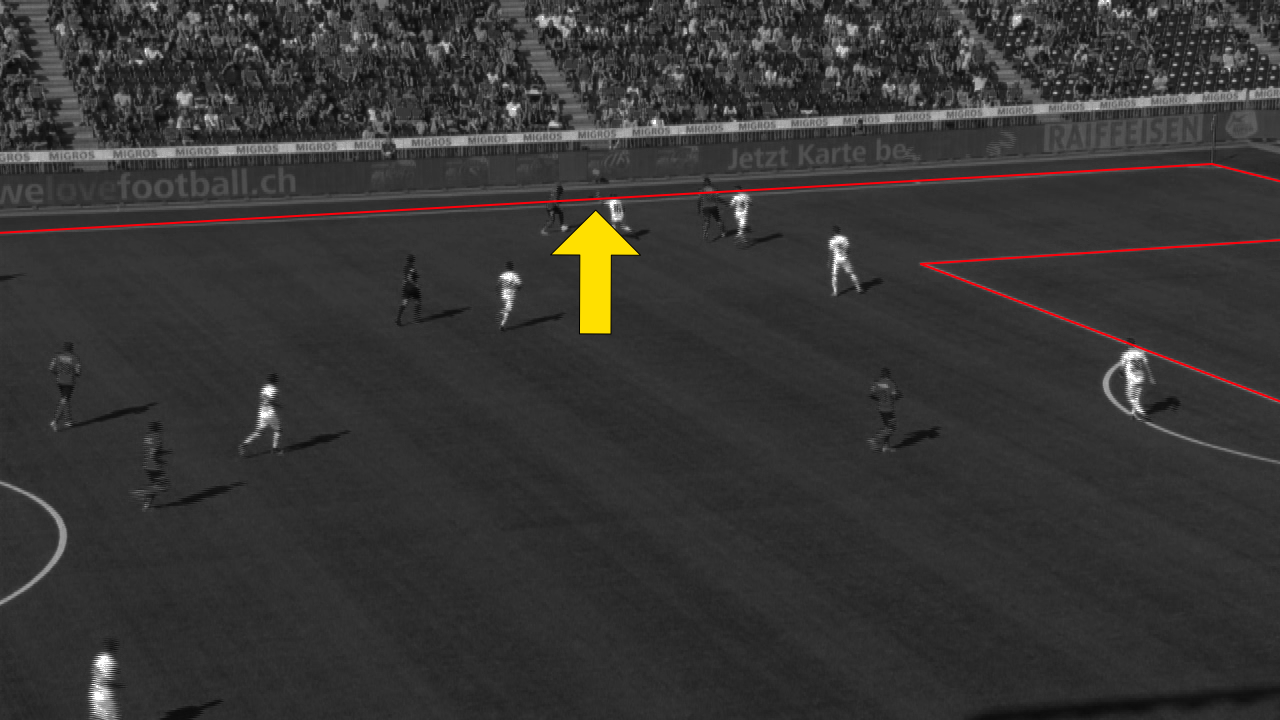}
    \includegraphics[width=6cm]{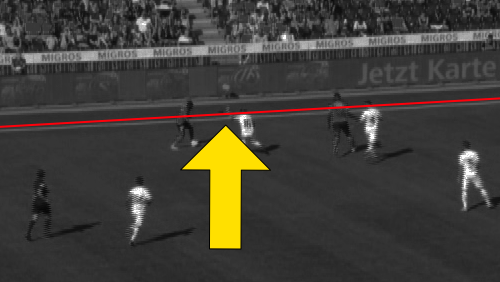}
    \includegraphics[width=6cm]{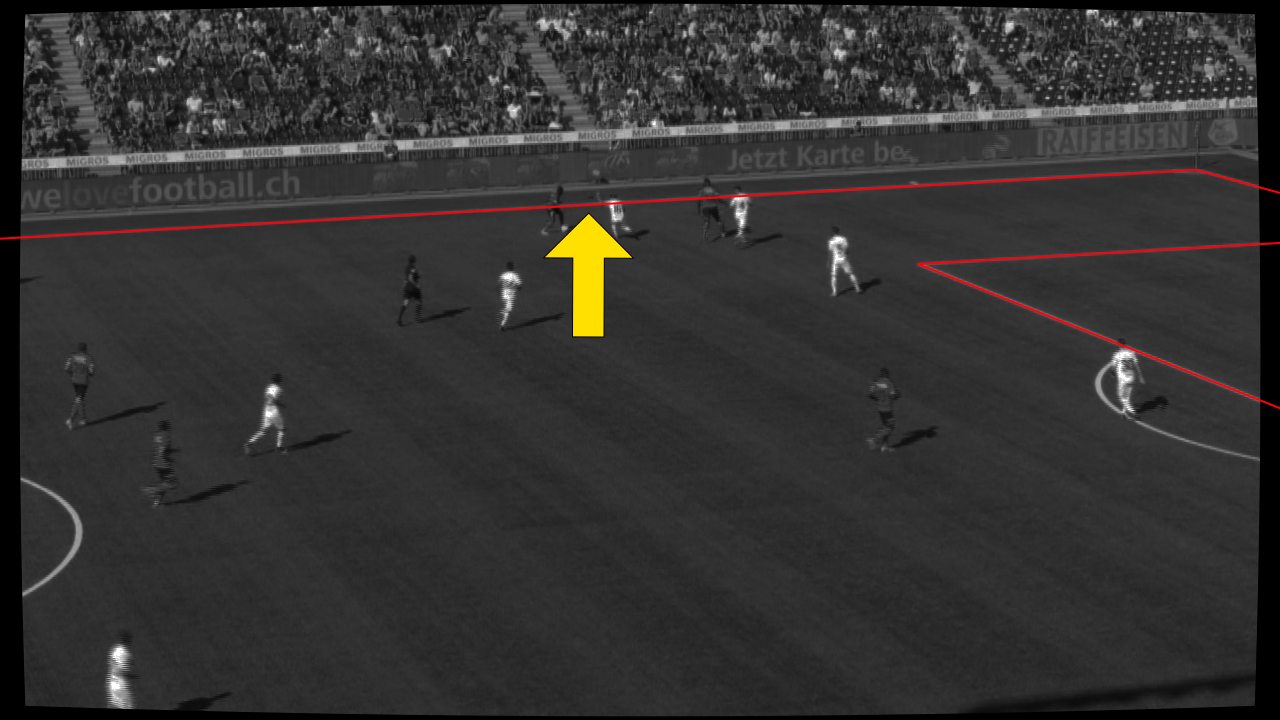}
    \includegraphics[width=6cm]{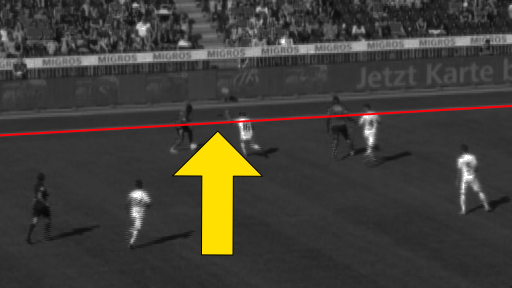}
    \caption{Original distorted image with red overlaid straight lines (top left), corrected image with red overlaid straight lines (bottom left), close-up of the distorted image where distortion is most prominent (top right), close-up of the corrected image where distortion is most prominent (bottom right).}
    \label{fig:my_label}
\end{figure}

\begin{figure}
    \centering
    \includegraphics[width=6cm]{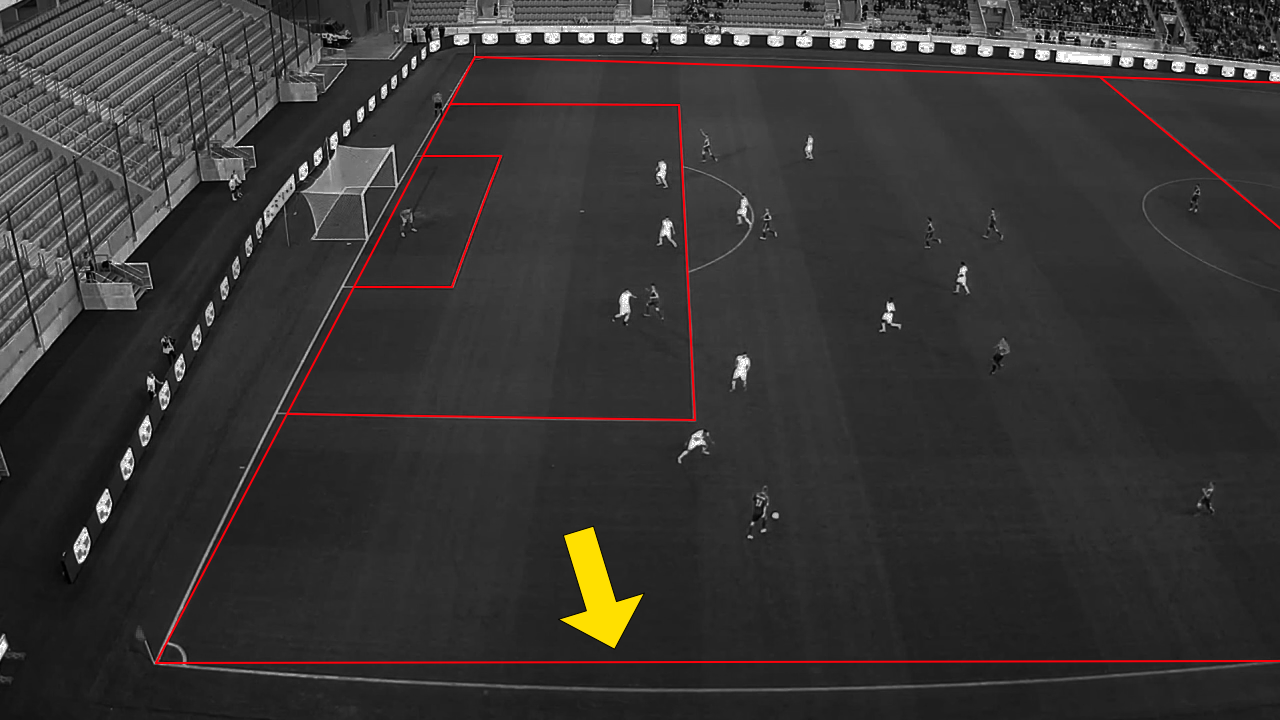}
    \includegraphics[width=6cm]{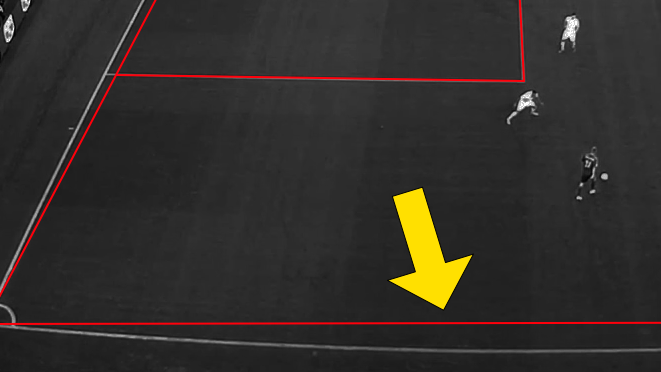}
    \includegraphics[width=6cm]{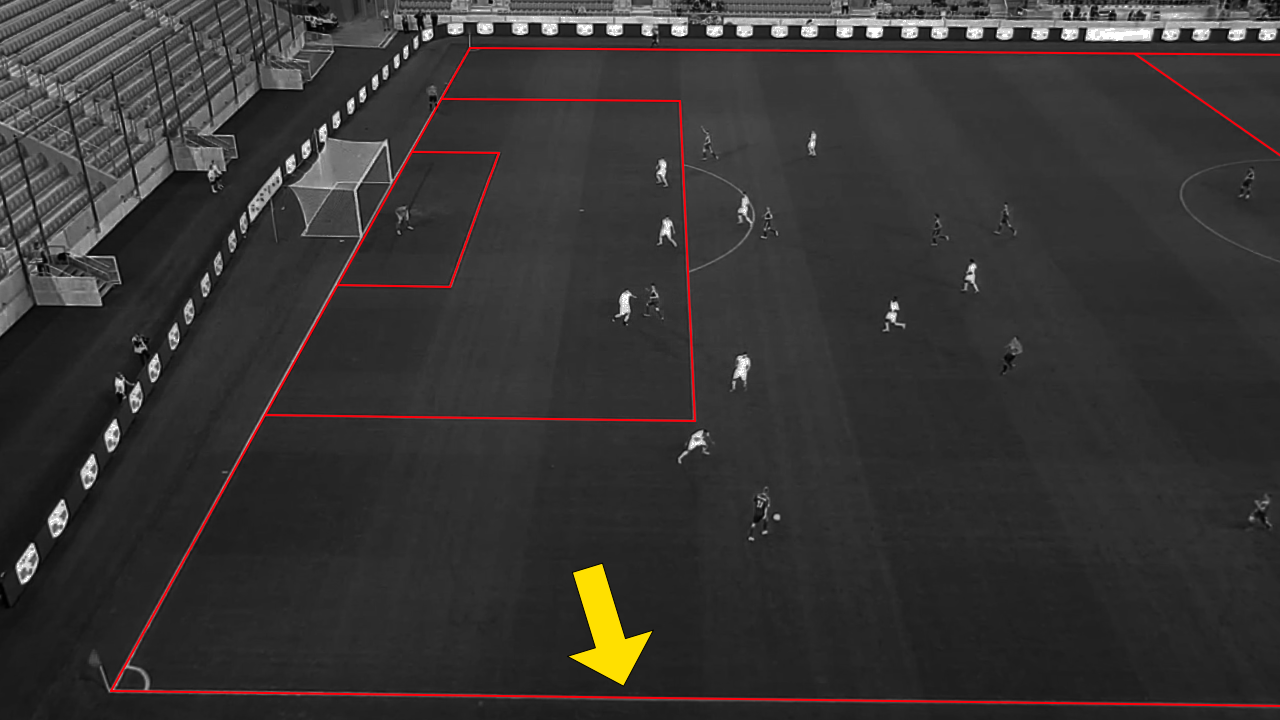}
    \includegraphics[width=6cm]{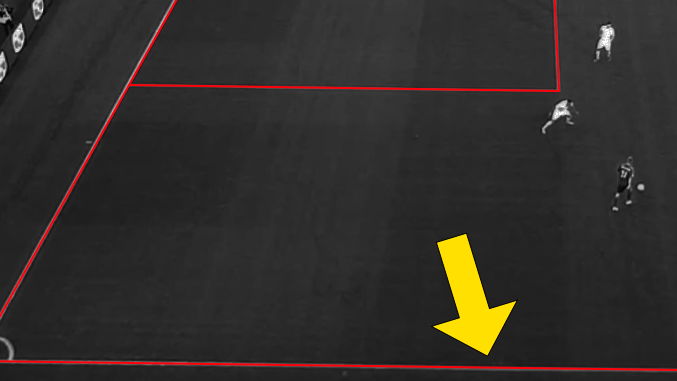}
    \caption{Original distorted image with red overlaid straight lines (top left), corrected image with red overlaid straight lines (bottom left), close-up of the distorted image where distortion is most prominent (top right), close-up of the corrected image where distortion is most prominent (bottom right).}
    \label{fig:my_label}
\end{figure}

\begin{figure}
    \centering
    \includegraphics[width=6cm]{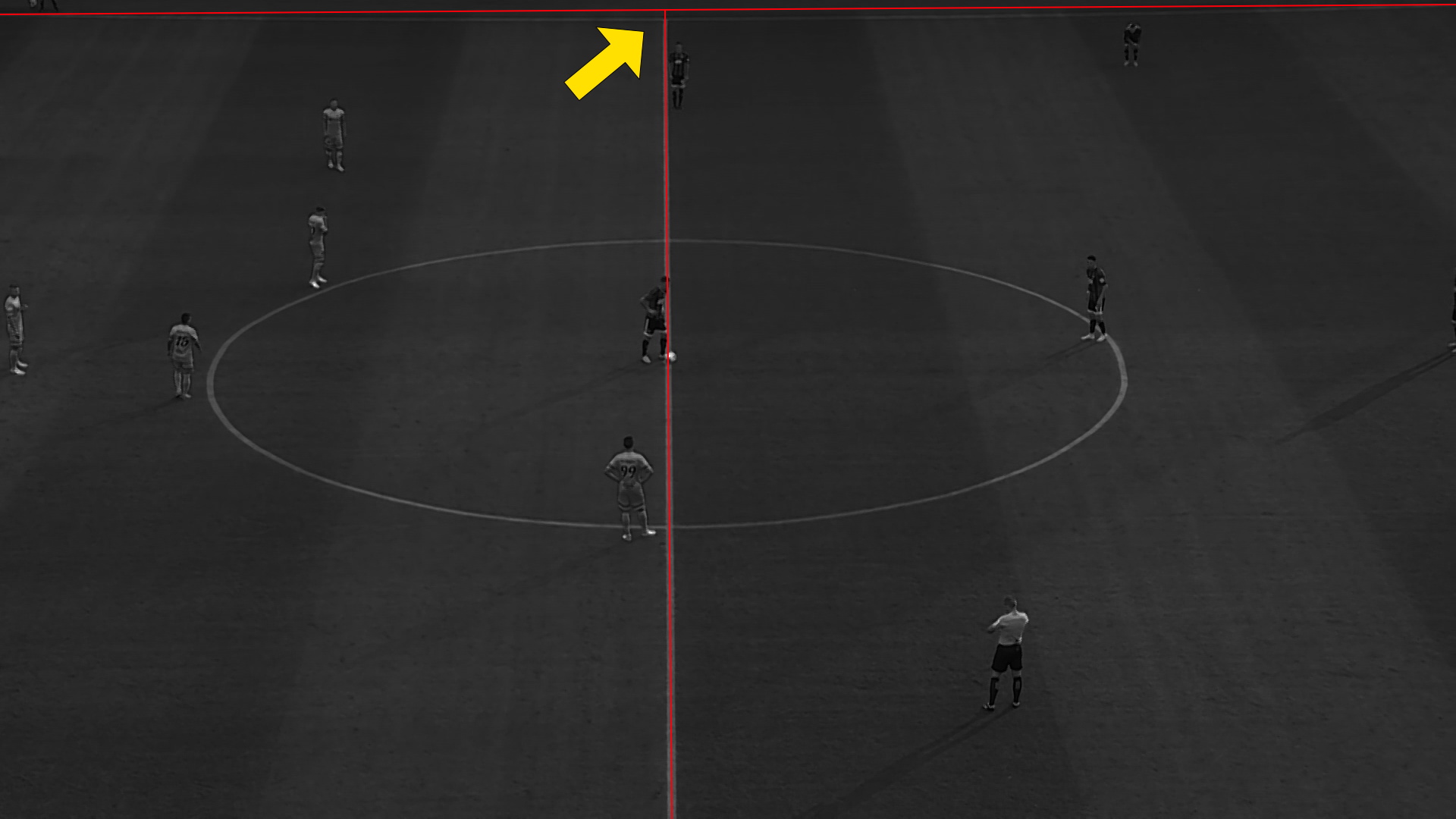}
    \includegraphics[width=6cm]{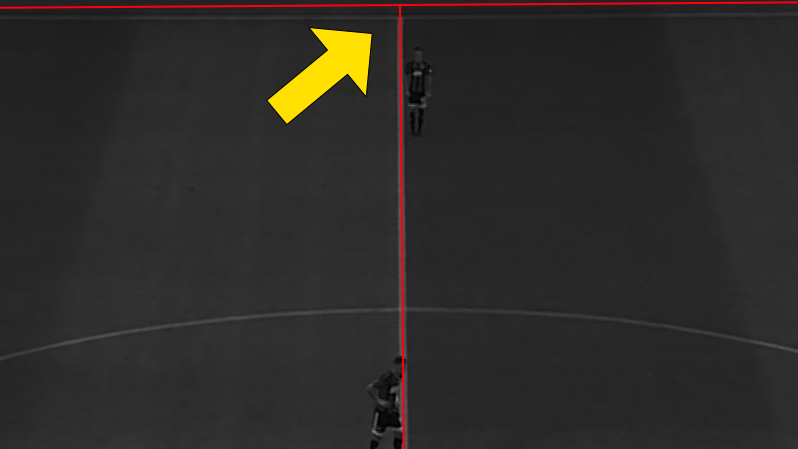}
    \includegraphics[width=6cm]{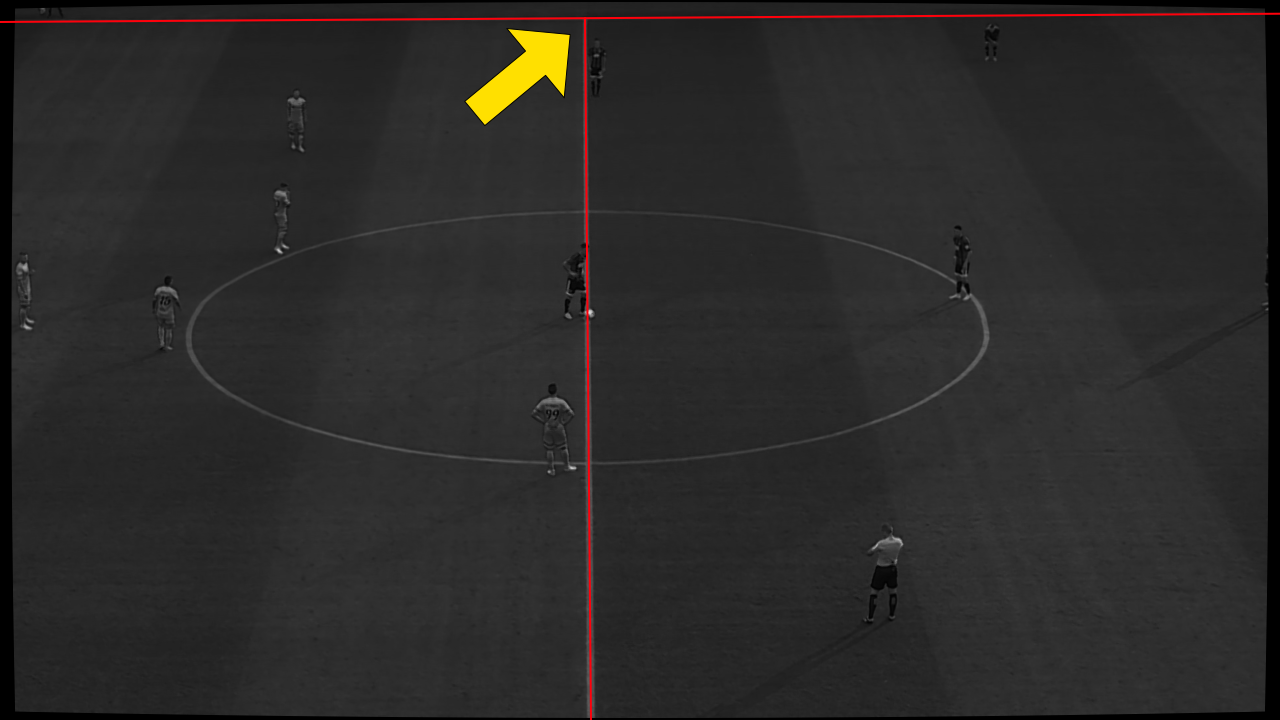}
    \includegraphics[width=6cm]{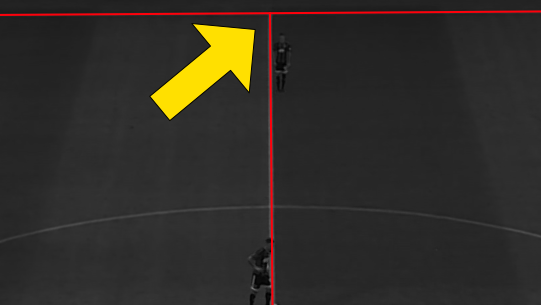}
    \caption{Original distorted image with red overlaid straight lines (top left), corrected image with red overlaid straight lines (bottom left), close-up of the distorted image where distortion is most prominent (top right), close-up of the corrected image where distortion is most prominent (bottom right).}
    \label{fig:my_label}
\end{figure}

\begin{figure}
    \centering
    \includegraphics[width=6cm]{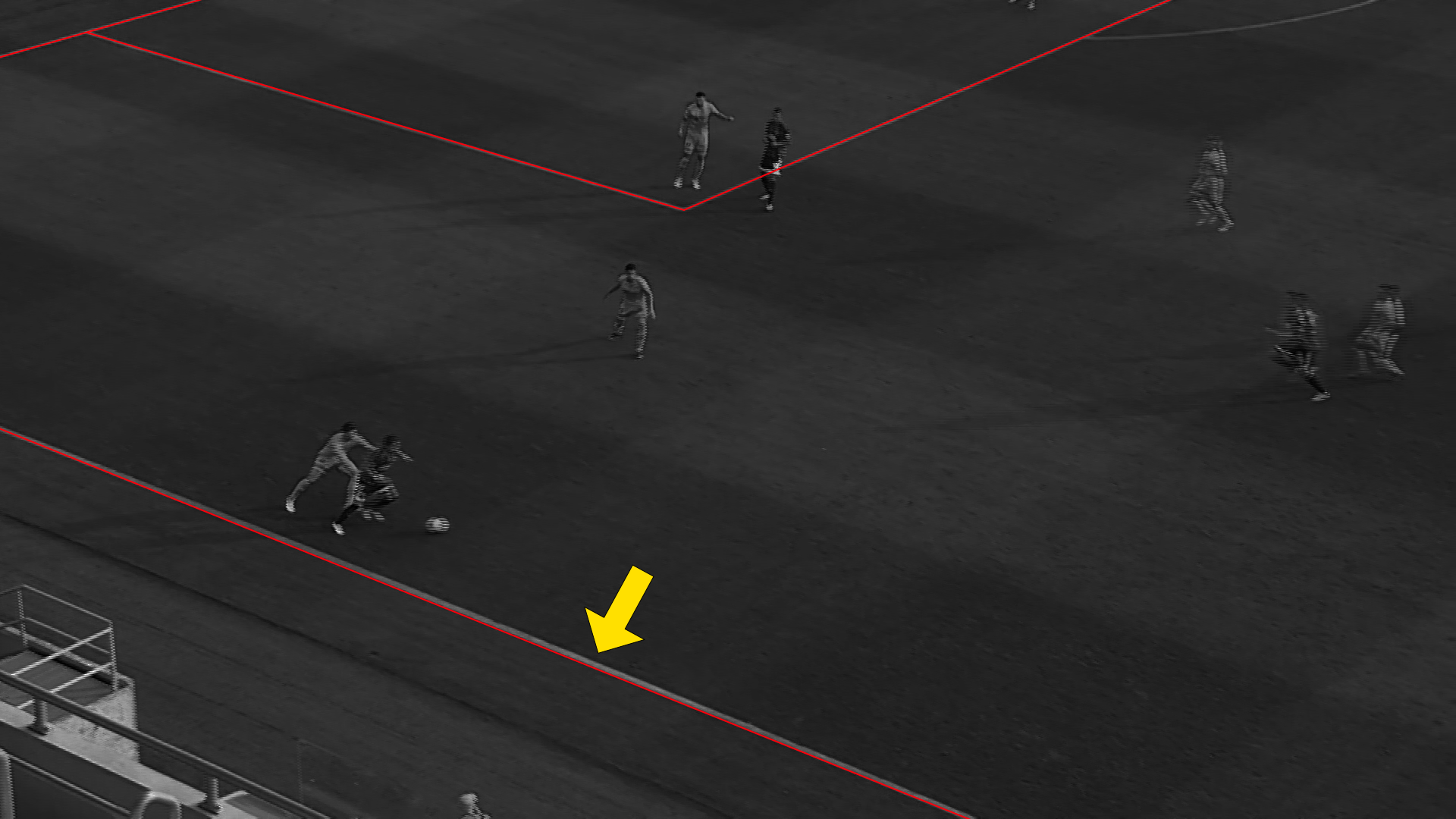}
    \includegraphics[width=6cm]{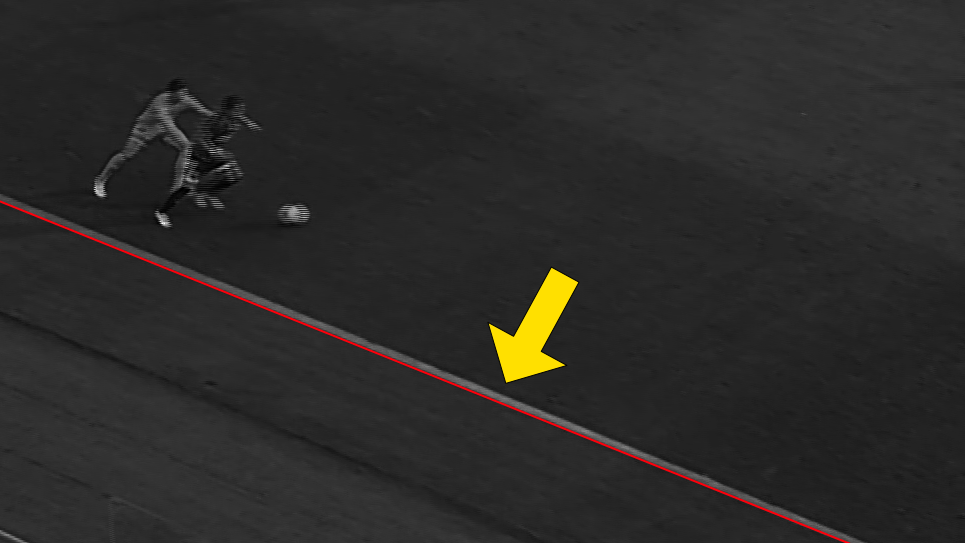}
    \includegraphics[width=6cm]{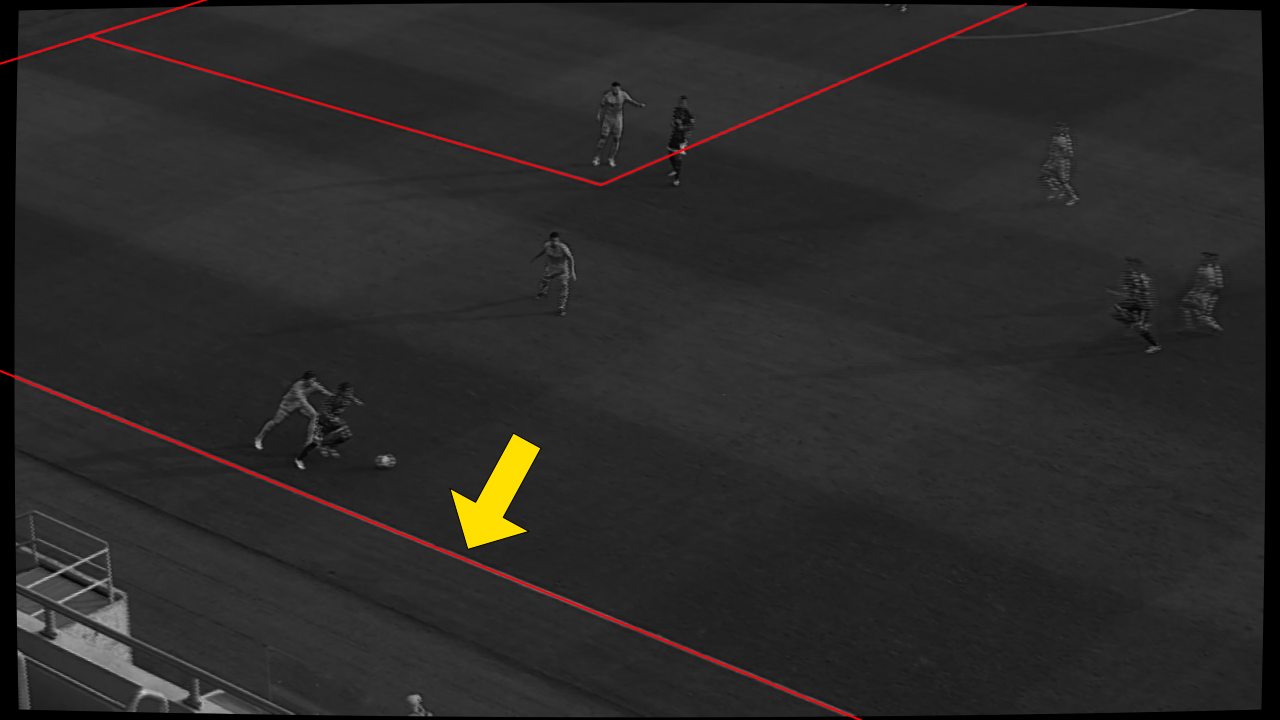}
    \includegraphics[width=6cm]{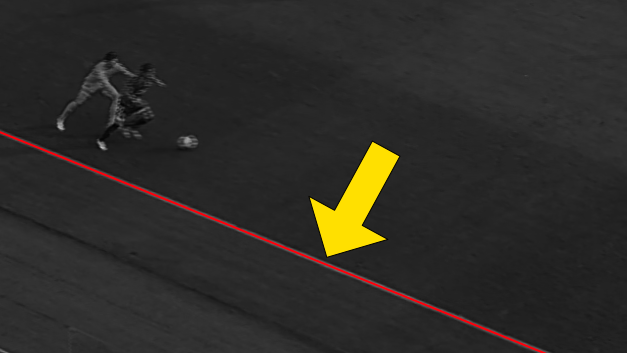}
    \caption{Original distorted image with red overlaid straight lines (top left), corrected image with red overlaid straight lines (bottom left), close-up of the distorted image where distortion is most prominent (top right), close-up of the corrected image where distortion is most prominent (bottom right).}
    \label{fig:my_label}
\end{figure}

\newpage
\bibliography{elsarticle-Janos}

\end{document}